# Research on Autonomous Robots Navigation based on Reinforcement Learning

Zixiang Wang*
College of Engineering and Computer Science
Syracuse University
Syracuse, NY
*Corresponding author e-mail address: zwang161@syr.edu

Hao Yan
College of Engineering and Computer Science
Syracuse University
Syracuse, NY
hyan17@syr.edu

Yining Wang
Human Factors in Information Design
Bentley University
Waltham, MA
wang_yini@bentley.edu

Zhengjia Xu
SATM
Cranfield University
Cranfield, UK
Billy.Xu@cranfield.ac.uk

Zhuoyue Wang
Department of Electrical Engineering and Computer Sciences
University of California, Berkeley
Berkeley, CA
zhuoyue_wang@berkeley.edu

Zhizhong Wu
Independent Researcher
Mountain View, CA
ecthelion.w@gmail.com

*Abstract*—Reinforcement learning continuously optimizes decision-making based on real-time feedback reward signals through continuous interaction with the environment, demonstrating strong adaptive and self-learning capabilities. In recent years, it has become one of the key methods to achieve autonomous navigation of robots. In this work, an autonomous robot navigation method based on reinforcement learning is introduced. We use the Deep Q Network (DQN) and Proximal Policy Optimization (PPO) models to optimize the path planning and decision-making process through the continuous interaction between the robot and the environment, and the reward signals with real-time feedback. By combining the Q-value function with the deep neural network, deep Q network can handle high-dimensional state space, so as to realize path planning in complex environments. Proximal policy optimization is a strategy gradient-based method, which enables robots to explore and utilize environmental information more efficiently by optimizing policy functions. These methods not only improve the robot's navigation ability in the unknown environment, but also enhance its adaptive and self-learning capabilities. Through multiple training and simulation experiments, we have verified the effectiveness and robustness of these models in various complex scenarios.

## I. INTRODUCTION

Mobile robots can make autonomous decisions and perform tasks according to changes in the environment, and their autonomous navigation capabilities are one of the key technologies to achieve industrial automation, improve production efficiency and reduce labor costs. In a complex and changeable production environment, path planning is the core technology for robots to achieve autonomous navigation. An excellent path planning algorithm can effectively reduce transportation costs, improve the efficiency of transportation operation management, and ensure the efficient operation of intelligent logistics systems, thereby further improving the efficiency and competitiveness of industrial production [1]. At present, most of the use scenarios of mobile robots are customized scenarios, such as independent planning and construction or targeted transformation of factories, which are relatively simple and helpful for effective path planning of robots.

Path planning is a key part of robot navigation and is of great significance. In the early days, path planning mainly used pre-programmed paths, that is, a certain number of guide lines or magnetic guide markers were pre-installed on the ground of the factory or warehouse, and mobile robots automatically navigated according to these guide lines or markers. This type of path planning is known as guideway path planning [2]. Since the number and position of the guide lines or markers are fixed, this method makes path planning less difficult and ensures that the mobile robot does not deviate from the predetermined path while driving.

Although leader line path planning is simple and easy, there are some drawbacks. First of all, pre-cabling requires a certain amount of cost and time, and at the same time, there are certain restrictions on the layout of the factory or warehouse. Second, guide wires or markers may need to be rerouted due to changes in the layout of the factory or warehouse, increasing maintenance costs [3]. Finally, the application scenarios of guide line path planning are relatively limited, and it is difficult to adapt to the path planning needs of different scenarios and complex environments.

In recent years, reinforcement learning has shown great potential in the field of autonomous robotics, especially for decision-making and control tasks in unknown environments. By interacting with its surroundings, the robot can learn the optimal strategy. However, to find this strategy, reinforcement learning often requires the robot to explore as many actions as possible in the environment [4]. However, in practice, these actions can be a safety hazard and can have serious consequences for the robot or the environment. For example, a robot may attempt to collide with an obstacle, and this collision may damage the robot, resulting in the failure of the entire learning process. Therefore, in the practical application of reinforcement learning, it is essential to improve the safety of learning [5].

The learning agent in reinforcement learning is called an agent, and its learning goal is to map states to actions in order to maximize digital reward signals. Unlike supervised learning, reinforcement learning usually deals with sequence data, which is difficult to meet the characteristics of independent and identical distribution [6]. At the same time, in the process of interaction between the agent and the environment, the reward in a certain state is not necessarily fixed, but may also be related to different moments, that is, the agent may receive a delayed reward [7]. Reinforcement learning is also different from unsupervised learning, and although finding the implicit structure between the data may be helpful for the agent's learning, it does not directly solve the problem of maximizing the reward signal.

Although reinforcement learning has successfully realized the automatic learning of complex behaviors, its learning process requires a large number of experiments, and the convergence is difficult to guarantee. In contrast, animals usually learn new tasks in just a few experiments, thanks to their prior knowledge of the world [8]. Therefore, scholars have made many explorations by drawing on the experience pool in deep reinforcement learning to achieve rapid re-learning of dynamic environments.

## II. RELATED WORK

In recent years, scholars have made many innovations and improvements based on such methods. Wu et al. [9] proposed a hybrid algorithm combining a beetle antenna search algorithm and an artificial potential field algorithm for real-time path planning. Experiments show that this method can not only generate better paths, but also have significant advantages in planning time. Kashyap et al. [10] experimented in a dynamic terrain consisting of multiple NAO robots and some static obstacles using a combination of dynamic window method (DWA) and teaching-based optimization technique (TLBO). The results show that the technology shows robustness and effectiveness in the path planning and obstacle avoidance process of single and multiple humanoid robots to cope with static and dynamic terrain.

Molinos et al. [11] proposed the dynamic obstacle window method and the dynamic obstacle tree window method, which ensure the normal operation of the robot in a dynamic environment by incorporating improvements such as robot speed to evaluate the stability of the planned path. In order to solve the problem that the traditional DWA algorithm only considers the obstacles on the trajectory, Saranrittichai et al. [12] proposed a regional dynamic window method (RDW), which modifies the objective function and considers the obstacles near the trajectory at the same time. The experimental results show that the robot can drive more safely when encountering near-orbit obstacles. Randhavane et al. [13] proposed a new pedestrian feature model to identify pedestrians based on the trajectory information of pedestrians in the navigation environment, so as to promote robot perception and navigation and avoid collisions with pedestrians. Experiments have shown that the robot is capable of performing socially aware navigation among dozens of pedestrians.

The existing methods are mainly improved on the basis of the traditional global path planning algorithm and local path planning algorithm, and the path planning performance of the algorithm is improved by increasing the path smoothness and reducing the path length. The research of scholars has effectively promoted the progress of robot path planning. However, path planning is not always possible to design in advance, as global environmental information is not always available a priori. Traditional algorithms often rely on map information to calculate the cost function of path planning, and when the environment changes dynamically, the route needs to be re-planned, resulting in low efficiency.

## III. METHODOLOGIES

Autonomous robot navigation is a complex task that requires robots to autonomously plan paths and avoid obstacles in unknown or dynamic environments. Reinforcement learning has shown great potential in this field as a trial-and-error learning method that can continuously optimize decision-making through interaction with the environment. This section will introduce specific model methods in autonomous robot navigation, including deep Q network and proximal policy optimization models. We summarize the main parameters in Table 1.

TABLE I. PRIMARY NOTIONS

| Symbols | Utilizations |
|---|---|
| $a$ | Action |
| $s$ | State |
| $\gamma$ | Discount factor |
| $r_t$ | Reward for time step |
| $\theta$ | Network parameter |
| $\pi_\theta(\cdot)$ | Policy function |
| $\tau$ | State-action sequence |
| $J(\theta)$ | Expected rewards |
| $\widehat{A_t}$ | Advantage function |
| $\epsilon$ | Clip threshold |

### A. Deep Q network

Deep Q networks are the result of the combination of Q-learning and deep neural networks, which are used to solve problems in high-dimensional state spaces. The basic principle of Q-learning is to learn a Q function that represents the expected cumulative reward for performing action $a$ in state $s$, which is expressed as Equation 1.

$$Q(s,a) = \mathrm{E}[\sum_{t=0}^{T} \gamma^t r_t | s_t = s, a_t = a] \qquad (1)$$

Where $\gamma$ is the discount factor, and $r_t$ is the reward for time step $t$. In deep Q networks, the Q function is approximated by a neural network with a parameter $\theta$. Use empirical playback and target network mechanisms to stabilize the training process. The experience replay stores the state, action, reward, and next state $(s, a, r, s')$ of the interaction in the experience pool, from which samples are randomly selected for training to reduce the correlation between samples. The target network introduces a target Q network with the parameter $\theta'$, and the parameters of $\theta$ are copied to $\theta'$ at fixed steps to keep the target value stable. The loss function is Equation 2.

$$L(\theta) = \mathrm{E}_{(s,a,r,s') \sim D}$$
$$\cdot [(r + \gamma max_{a'} Q(s', a'; \theta') - Q(s, a; \theta))^2] \quad (2)$$

*B. Proximal policy optimization*

Proximal policy optimization is a reinforcement learning method based on policy gradient, which aims to improve the stability and sample efficiency of the policy gradient method. The strategy gradient approach maximizes the expected reward by directly optimizing the $\pi_\theta(a|s)$ of the policy function, which is expressed as Equation 3.

$$J(\theta) = \mathrm{E}_{\tau \sim \pi_\theta} [\sum_{t=0}^{T} \gamma^t r_t] \quad (3)$$

Where $\tau$ represents the state-action sequence and $J(\theta)$ means expected rewards. Proximal policy optimization avoids instability caused by drastic updates by limiting the step size of each policy update. The core of proximal policy optimization is to introduce clipping methods to constrain the change of old and new strategies, which is manifested in the following Equation 4.

$$L^{CLIP}(\theta) = \mathrm{E}_t \cdot [\min(\frac{\pi_\theta(a_t|s_t)}{\pi_{\theta old}(a_t|s_t)} \widehat{A_t},$$
$$clip(\frac{\pi_\theta(a_t|s_t)}{\pi_{\theta old}(a_t|s_t)}, 1-\epsilon, 1+\epsilon)\widehat{A_t})] \quad (4)$$

Where $\widehat{A_t}$ is the advantage function and $\epsilon$ is the clip threshold to ensure that the policy change does not exceed a certain range. The loss function after clipping is $L^{CLIP}(\theta)$. Through the above methods, robots can achieve efficient and stable autonomous navigation in dynamic and complex environments, effectively improving the performance of industrial automation and intelligent logistics systems.

Proximal policy optimization is a reinforcement learning method based on policy gradient, which aims to improve the stability and sample efficiency of the policy gradient method. Proximal policy optimization maximizes the expected reward by directly optimizing the policy function and limits the step size on each policy update to avoid the instability caused by drastic updates. The core of this is to introduce the editing method to constrain the changes of the old and new strategies, ensure that the policy changes are within a reasonable range, so as to achieve a more stable training process.

## IV. PREPARE YOUR PAPER BEFORE STYLING

*A. Experimental setups*

In this experiment, we evaluated the performance of different reinforcement learning algorithms in robot navigation tasks. The robot needs to navigate a 10×10 grid world environment and find the best path to reach the target location. In order to compare the effectiveness of different algorithms, we made the following settings: the environment settings included a grid size of 10×10, the starting position in the lower left corner, and the target position in the upper right corner, and a random number of obstacles, the position remained the same in each experiment. Following Figure 1 shows the simulation environment.

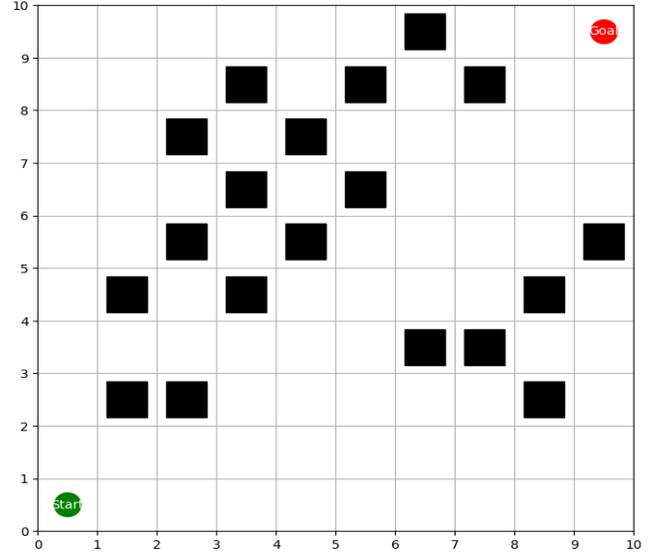

Figure 1. 10x10 Grid World Environment.

We used different reinforcement learning algorithms, including Deep Q Network (DQN) and Proximal Policy Optimization (PPO), each algorithm conducted 100 independent experiments, each experiment contained 500 episodes, and the number of collisions in each independent experiment was recorded. The evaluation index is the number of collisions, and the performance is evaluated by comparing the average number of collisions of different algorithms 100 independent experiment.

*B. Experimental analysis*

We compare the performance of different reinforcement learning algorithms by plotting robot navigation paths, and in order to visually compare the performance of different algorithms in robot navigation tasks, we plot the navigation paths generated by each algorithm. Following Figure 2 shows the different navigation paths for different methods.

The number of collisions is an important indicator to evaluate the performance of autonomous robot navigation algorithms, which directly reflects the safety and obstacle avoidance ability of robots in the navigation process. By conducting multiple independent experiments in a 10×10 grid environment, the number of collisions of each algorithm in 100 rounds of episodes was recorded, and the average value was calculated to evaluate its performance. A lower number of collisions means better obstacle avoidance and higher navigation efficiency. By plotting a line graph of the number of collisions, you can visually compare the long-term stability and robustness of different algorithms.

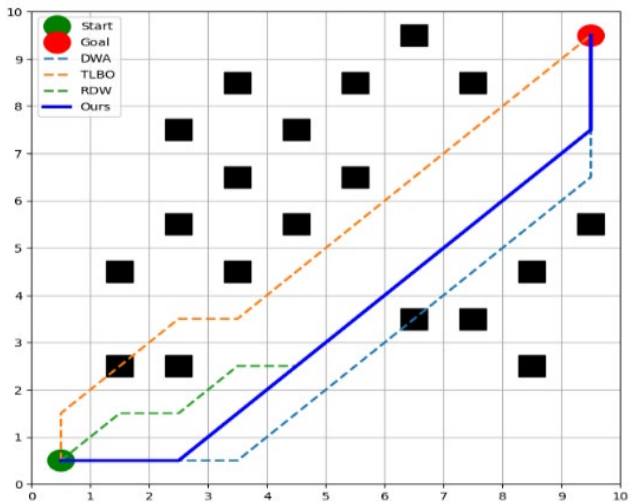

Figure 2. 10x10 Grid World Environment with Navigation Paths.

Our experimental results show that our method outperforms other methods in terms of the number of collisions, demonstrating higher navigation efficiency and safety. Following Figure 3 compares the collisions results.

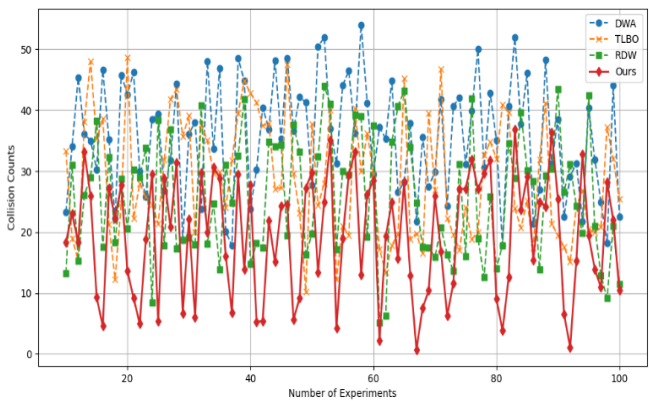

Figure 3. Comparison of Collision Counts in Different Experiments.

Path smoothness is another important indicators to evaluate the navigation performance of robots. It measures the continuity and smoothness of the robot's path. Specifically, fewer sharp turns and discontinuities in the path indicates a smoother path. A smooth path not only reduces the robot's energy consumption and wear and tear in motion, but also improves its overall efficiency and operating life. In addition, the smooth path helps reduce impact and stress on mechanical components, reduces maintenance costs, and improves the reliability of the navigation system. Figure 4 shows the path smoothness comparison results.

## V. Conclusions

In conclusion, our research on autonomous robots navigation based on reinforcement learning demonstrates the effectiveness and potential of advanced RL algorithms in improving robotic navigation performance. Our proposed method in a 10×10 grid world environment, we highlighted key metrics like collision counts and path smoothness. Our findings indicate that our method consistently outperforms others, showcasing superior safety, efficiency, and adaptability. Enhanced path smoothness not only reduces energy consumption and wear but also contributes to longer operational life and reduced maintenance costs. These results underscore the importance of reinforcement learning in developing robust and efficient autonomous navigation systems, paving the way for future advancements in industrial automation and intelligent logistics.

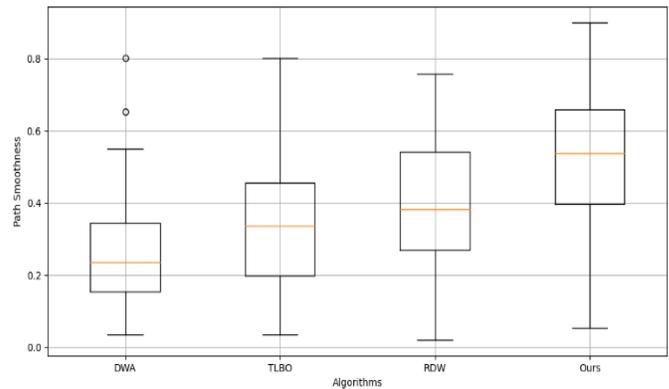

Figure 4. Comparison of Path Smoothness in Different Algorithms.